# Bio-inspired Robustness: A Review

Machiraju, Harshitha[1,2]* Choung, Oh-Hyeon[1]* Frossard, Pascal[2] , Herzog, Michael. H.[1]

[1]Laboratory of Psychophysics, Brain Mind Institute, École Polytechnique Fédérale de Lausanne (EPFL), Switzerland, http://lpsy.epfl.ch, harshitha.machiraju@epfl.ch, oh-hyeon.choung@epfl.ch

[2]Signal Processing Laboratory 4 (LTS4), École Polytechnique Fédérale de Lausanne (EPFL), Switzerland, https://www.epfl.ch/labs/lts4/

* The authors contributed equally.

## Abstract

Deep convolutional neural networks (DCNNs) have revolutionized computer vision and are often advocated as good models of the human visual system. However, there are currently many shortcomings of DCNNs, which preclude them as a model of human vision. For example, in the case of adversarial attacks, where adding small amounts of noise to an image, including an object, can lead to strong misclassification of that object. But for humans, the noise is often invisible. If vulnerability to adversarial noise cannot be fixed, DCNNs cannot be taken as serious models of human vision. Many studies have tried to add features of the human visual system to DCNNs to make them robust against adversarial attacks. However, it is not fully clear whether human vision inspired components increase robustness because performance evaluations of these novel components in DCNNs are often inconclusive. We propose a set of criteria for proper evaluation and analyze different models according to these criteria. We finally sketch future efforts to make DCCNs one step closer to the model of human vision.

# 1. Introduction

Deep convolutional neural networks (DCNN) have revolutionized computer vision. DCNNs reach near or even super-human performance in many tasks, such as image classification (He et al., 2016), image segmentation (He et al., 2017), image captioning (Karpathy & Fei-Fei, 2015), and image generation (Choi et al., 2020). There is now a fierce debate whether DCNNs are also a good model for the human visual system. On the one hand, proponents argue that DCNNs perform like humans in many object recognition tasks, and their architecture indeed resembles the human one (Kubilius et al., 2018; Schrimpf et al., 2020). On the other hand, DCNNs often solve vision tasks very differently than humans (Geirhos et al., 2020), indicating that comparable performance levels *per se* do not tell whether DCNNs are good models.

In this contribution, we look at this pro-con dichotomy from the perspective of robustness, that is, the ability of DCNNs to make proper object classifications even when data are slightly perturbed or coupled with noise. Specifically, we consider adversarial attacks, which present a major problem for DCNNs (Sharif et al., 2016; Y. Zhang et al., 2019). These attacks are small, crafted perturbations that cause major misclassifications of the DCNNs even though they are imperceptible to humans (Szegedy et al., 2014). Figure 1 shows two adversarial examples where the image of a dog with imperceptible noise is misclassified as red wine or toilet paper by DCNNs. Obviously, humans and DCNNs show very different behavior, hence at first glance, DCNNs should be discarded as proper models for human vision. However, it may be that minor fixes can make DCNNs robust to adversarial attacks, for example, by taking inspiration from human vision models. Such attempts may help to bridge the performance gap that still exists between DCNNs and human vision.

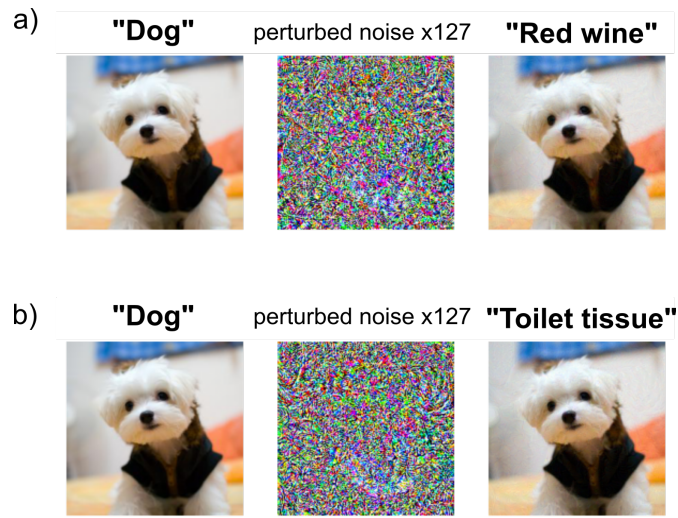

*Figure 1. Adversarial examples, using PGD with $L_\infty$ and with noise constraint of $\epsilon = 16/255$ on 'n02085936_6883.jpeg' of ImageNet (Deng et al., 2009) dataset. The left image is the original, and the right image is the original plus the noise image shown in the middle. For humans, the differences between the original image and the original plus noise image are hardly visible. For DCNNs, the noise leads to serious misclassification.*

Many investigations have proposed improvements against adversarial attacks from both the Computer vision (e.g., Athalye et al., 2018b; Carlini et al., 2019; Croce et al., 2020; Tramer et al., 2020) and the Neuroscience communities (e.g., Choksi et al., 2020; Kiritani & Ono, 2020; Marchisio et al., 2020; Rusak et al., 2020; Zoran et al., 2020). Improving robustness is crucial not only to reduce vulnerability to adversarial attacks but also for improving the transfer of learning (Salman et al., 2020; Utrera et al., 2020), image segmentation (Salman et al., 2020), generalization (Bochkovskiy et al., 2020; Song et al., 2020; Xie et al., 2020), etc. (see Fig. 2). In this paper, we focus on bio-inspired methods, with the main objective to establish stronger connections between DCNNs and human vision.

Unfortunately enough, different bio-inspired approaches trying to protect DCNNs against adversarial attacks by including components of the biological visual system have often reached unresolved conclusions. We advocate that one of the main reasons for this situation is the non-uniform ways of evaluating and analyzing these components.

In this paper, we try to remedy this situation by proposing criteria to standardize the evaluation and analysis methods that each study needs to meet. We first review definitions of robustness and adversarial attacks, and evaluation methods in Section 2. We then explore different studies and summarize their results and analysis in Sec 3 and 4. Finally, we summarize the main learning messages and propose new insights for future research directions towards understanding better how the joint development of DCNN architectures and human vision models could lead to cross-fertilization, which could for example lead to better and more robust computer vision systems.

## Robustness

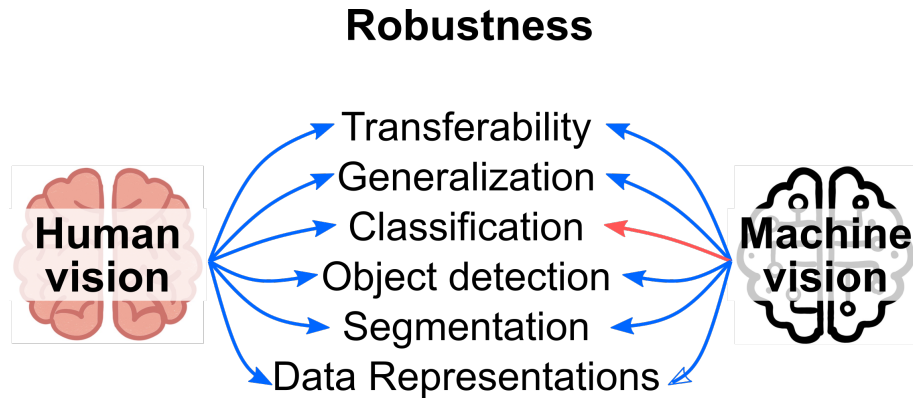

*Figure 2. Robust features are transferable to other tasks. Since most Deep Learning models are solely optimized to be highly accurate for a given task, they do not generalize to other tasks, unlike humans (Lapuschkin et al., 2019). Ilyas et al., 2019 showed that by design, Deep Learning models tend to use highly predictive, non-robust features (which include background, texture/high-frequency information, etc.) instead of robust features (which include foreground, shape/low-frequency information; Zhang & Zhu, 2019, etc.). This explains, to some extent, the high vulnerability of DCNNs to adversarial attacks. Hence, by forcing DCNNs to utilize robust features, we may obtain more human-vision-like representations (Engstrom et al., 2019; Kaur et al., 2019; Tsipras et al., 2018). Since robust networks utilize global information like shapes, they have a better "understanding" of the overall features of each class, which improves generalization to unseen distributions (Bochkovskiy et al., 2020; C. Song et al., 2020; Cihang Xie et al., 2020), better image segmentation and object detection abilities (Salman et al., 2020), and adaptation across domains since the features learned are now generic for each class (transfer learning; Salman et al., 2020; Utrera et al., 2020).*

## 2. Adversarial Robustness

### 2.1. Adversarial Attacks and Defenses

In adversarial attacks, carefully chosen noise added to images containing objects leads to gross misclassification of these objects (Szegedy et al., 2014, Fig. 1). Interestingly, the noise is usually imperceptible for humans. For example in Fig .1, humans can hardly distinguish the original image with the dog and the same image with additive noise. Adversarial attacks are mathematically defined as follows: for a given image $x$, and a classifier $f: \mathbb{R}^n \to \mathbb{R}^c$, where $x \in \mathbb{R}^n$ and $y$ is the **true** class label we define the adversarial noise $\Delta x$ as:

$$arg\ max\ (f(x + \Delta x)) \neq arg\ max\ (f(x))\ \text{s.t}\ ||\ \Delta x\ ||_p \leq \epsilon \qquad \text{(Eq. 1)}$$

The true label class ($y$) that the input ($x$) belongs to is referred to as the *source class,* and the *target class* is the class to which the adversarial sample ($x + \Delta x$) is misclassified by the DCNN model. In the above equation, the upper bound on the noise $\epsilon$ is added to ensure that the noise remains small.

In Eq. 1, Adversarial perturbations are defined w.r.t. an $L_p$ norm. The usual values for *p* are *0, 1, 2, and* $\infty$ can be summarized as follows:

- $L_0$ norm: Constrains the number of non-zero pixels in the adversarial perturbation (sparsity).
- $L_1$ norm (Manhattan distance): Constrains the absolute value of the magnitude of the adversarial perturbation.
- $L_2$ norm (Euclidean distance): Constrains the squared value of the magnitude of the adversarial perturbation.
- $L_\infty$ norm: Constrains the maximum element of the adversarial perturbation.

The upper bound for adversarial perturbations, $\epsilon$ is chosen w.r.t. an $L_p$ norm (Fezza et al., 2019; Jordan et al., 2019) such that the noise is imperceptible to humans (Bouniot et al., 2021).

Adversarial perturbations in Eq. 1 may be defined w.r.t. any norm or distance metrics other than the $L_p$ norm metrics (Jordan et al., 2019). But, the $L_p$ metrics are most studied, well understood, and are much simpler to use than other distance metrics (Akhtar & Mian, 2018; Ortiz-Jimenez et al., 2020). Hence throughout this paper, we focus on $L_p$ norm based adversarial attacks and robustness.

There are different algorithms to generate adversarial attacks, including the Fast Sign Gradient Method (FGSM; Szegdy et al., 2014), Projected gradient descent (PGD; Madry et al., 2017), Jacobian Based Saliency Map Attacks (JSMA; Papernot et al., 2016), DeepFool (Moosavi-Dezfooli et al., 2016), Decision boundary attacks (DBA; Brendel et al., 2018), Carlini-Wagner attacks (Carlini & Wagner, 2017), Basic iterative method (BIM; Kurakin et al., 2017), etc. More information can be found in reviews such as Akhtar & Mian, 2018 and Ortiz-Jimenez et al., 2020.

DCNNs, unlike humans, do not have any understanding of object features. DCNNs learn discriminative features that are maximally capable of providing high accuracy on the trained data and construct decision boundaries based on these features. These learned features focus on local information of the image, such as texture (Song et al., 2020). It is this over-reliance on local features that makes DCNNs an easy prey for small adversarial noises (Ilyas et al., 2019).

The machine learning community has explored different ways to make DCNNs more robust (defend) against adversarial perturbations. Adversarial training (Madry et al., 2017; Moosavi-Dezfooli et al., 2016) is regarded as the state-of-the-art defense method. This method trains DCNNs by utilizing adversarial samples generated by attacks and augmenting the training dataset with these samples. By doing so, the DCNN is forced to change its decision boundaries to make sure that the adversarial samples belong to the same class as the original input sample. However, this type of training comes with a large computational load. For each input sample, the DCNN must first create adversarial perturbations via attacks and then learn to classify them correctly. Considering humans do not have to follow a similar procedure, there might be a simpler way to make DCNNs robust to adversarial perturbations. Many works have tried to use human vision-like features for creating more robust DCNNs (Chakraborty et al., 2018). These studies mainly focus on changing the decision boundaries of the DCNNs by learning global rather than local features similar to human vision (Engstrom et al., 2019).

However, as shown by Carlini et al., 2019 due to the failure to systematically evaluate the robustness of a defense, it is unclear how good these methods are (Brendel & Bethge, 2017 vs. Nayebi & Ganguli, 2017).

## 2.2. Evaluation of Robustness

### 2.2.1 Evaluation

There are different ways to measure the robustness of a DCNN against adversarial attacks, including:

**Adversarial Accuracy**: Defined as the fraction of correctly classified adversarial samples out of all adversarial samples created on a dataset:

$$\text{Adversarial Accuracy} = \frac{\#correctly\ classified\ adversarial\ samples}{\#adversarial\ samples} \qquad \text{(Eq. 2)}$$

Most work uses this metric to measure robustness.

**Attack success rate:** Defined as the fraction of adversarial samples generated by a given attack that succeeds in fooling the DCNN model out of all the generated adversarial samples.

$$\text{Attack success rate} = \frac{\#misclassified\ adversarial\ samples}{\#adversarial\ samples} \qquad \text{(Eq. 3)}$$

**Mean Distortion:** Defined as the median distance of all pairs built on a source image in a given dataset ($D$) and the corresponding adversarial sample.

$$\text{Mean distortion} = median\ (\sum_{\forall\ x\ \in\ D} ||\ x_{adv} - x\ ||_p\ )\ \text{where } p \in \{0,\ 1,\ 2, \infty\} \qquad \text{(Eq. 4)}$$

In general, the evaluation of robustness heavily depends on the attack used for generating the adversarial samples. The adversarial attack itself depends on parameters such as the number of iterations, the $\epsilon$ values, step sizes, etc. Typically, as many adversarial attacks are constructed on the estimation of gradients in Eq. 1, one has to make sure that the evaluation really captures the intrinsic robustness of the model, and not merely the impossibility to compute adversarial examples through specific, gradient-based methods, referred to as *gradient masking*.

DCNNs learn through backpropagation and minimize the classification loss by using the gradient of the loss function. Adversarial perturbations are created by maximizing the classification loss to cause misclassification. Thus, adversarial perturbations need gradients to maximize the classification loss.

Gradient masking refers to hiding this gradient information by using non-differentiable operations like subsampling (Hosseini et al., 2019), input transformations (Guo et al., 2018), or stochasticity (Dhillon et al., 2018). Gradient masking does not lead to adversarial robustness, it only makes it difficult for gradient-based adversarial attacks to find adversaries (Athalye et al., 2018). If an attack succeeds in finding a hidden adversarial sample (despite gradient masking), the DCNN still ends up misclassifying it. Gradient masking does not improve the decision boundary, it just avoids existing gradient-based adversarial attacks.

Given the high variability of methods for the evaluation of adversarial robustness, for any defense algorithm that aims at improving robustness, it is necessary to follow the basic defense evaluation guidelines provided by Carlini et al., 2019. Specifically, we stress the importance of checking for gradient masking phenomena, as explained above.

## 2.2.2 Bio-inspired robustness evaluation

As explained earlier, if bio-inspired defenses use gradient masking, they are not truly making DCNNs functioning closer to human visual perception. Hence, bio-inspired robustness methods need to ensure that they are not using gradient masking. Here, we summarize the sanity checks to identify gradient masking (for 1-4 Athalye et al., 2018 ):

1. **One-step attacks perform better than iterative attacks:** When an adversarial attack comes with multiple iterations (Madry et al., 2017), it has a better chance to succeed than an attack with a single iteration (Szegedy et al., 2014), e.g., PGD (multi-iteration attack) vs. FGSM (1 iteration attack). Failure to see this usually indicates gradient masking.

2. **Observing irregularities in perturbation budget curves:** Perturbation budget refers to the maximum value of $\epsilon$ used to generate adversarial examples as in Eq. 1. The higher the perturbation budget, the better the chance of attack success. Plots of the Attack success rate vs. perturbation budget are called perturbation budget curves. Ideally, the attack success rate should increase with the perturbation budget (larger noise) and for extremely high values of $\epsilon$ the attack success rate would be 100%. If this behavior is not seen in the perturbation curve, there is likely gradient masking.

3. **Random sampling finds adversarial perturbations:** Adversarial attacks are optimized to find adversarial perturbations and hence have a much higher chance of causing misclassification than any randomly sampled perturbation. If the success rate of adversarial attack is lower than random perturbations there is likely gradient masking.

4. **Black-box attacks are better than white-box attacks:** In order to generate adversarial perturbations, most attacks typically need information about the model being used, especially its parameters. When attacks are generated with complete access to model information, they are called white-box attacks. Attacks without any access to the model and its parameters are called black-box attacks. Black-box attacks usually have a lower attack success rate than white-box attacks. Failure to see this usually indicates gradient masking.

5. **Non-gradient-based adversarial attacks:** Other than the above suggested by Athalye et al., 2018, non-gradient-based adversarial attacks like Decision Boundary attacks (DBA; Brendel et al., 2018) can also be used to identify gradient masking. Let us consider the robustness of a standard model (without any defense method) evaluated on a non-gradient-based attack to be $r_s$, and for the defended model the robustness is $r_d$. If the defense uses gradient masking, it will be broken by the non-gradient-based attack, since masked (hidden) gradient information makes no difference to such attacks. Hence, for defenses with gradient masking, $r_d \approx r_s$. If there is no gradient masking $r_d > r_s$.

Since the use of non-differentiable layers or stochasticity could result in gradient masking, the above checklist can be used to validate the evaluation of the actual robustness of a model. Additionally, the adversarial attacks proposed by Athalye et al., 2018, which are specially designed to break gradient masking based defenses, can also be used to evaluate defenses with similar components.

# 3. Biologically Inspired Components

We review a series of methods, which propose to increase robustness against adversarial attacks by adding features of the human visual system to DCNNs. We first review some design guidelines, and then we highlight specific bio-inspired components and their properties. These methods were selected from the recent machine learning literature and are summarized in Table 1.

## 3.1. Choice of Bio-Inspired Components

In bio-inspired robustness, the idea is to make DCNN robust like humans while employing the concepts used by the latter. One problem is that it is rather unclear how the visual system acquires robustness. Hence, when implementing a new component, it is important to motivate, analyze, and validate the suggested components. Accordingly, when applying bio-inspired components for robustness, evidence should be provided that these components actually improve the DCNN.

In particular, it is necessary to test whether the contribution of each component indeed increases robustness. We can do so by "freezing" the component and analyze the change in robustness in comparison when the component is active. Alternatively, we can use simpler variants of the bio-inspired component.

Finally, it is also important to validate the design choices by verifying that the proposed bio-inspired component is indeed increasing robustness for the reasons it was motivated. For example, if Gaussian filtering is thought to improve robustness because it filters out adversarial noises, the authors should show that indeed adversarial noise is filtered out, on top of other robustness measures (Xie et al., 2019). Methods for validation include weight visualization, representation visualization, saliency maps, etc.

We review below different recent studies, which have proposed to include bio-inspired components in order to improve the robustness of DCNNs, and discuss motivations, design choices, and expected benefits.

## 3.2. Early visual processes

In human vision, the visual inputs are processed in the following order:

*Retina → Lateral Geniculate thalamic Nucleus (LGN) → Primary Visual Cortex (V1)*

The retina has a higher density of photoreceptors in the fovea than in the periphery. This uneven distribution of photoreceptors results in a non-linear sampling of the visual input. Therefore, the image resolution is increased in the fovea and blurred towards the periphery. This, in turn, increases the Signal-to-Noise Ratio (SNR) of the input, which was proposed to increase robustness (Elsayed et al., 2018).

Retinal neurons use lateral inhibition, which attenuates redundant and irrelevant signals (decorrelation; Segal et al., 2015). Thus, only significant signals (Bakshi & Ghosh, 2017), which are required to properly classify the object, survive. Thus, we suggest lateral inhibition may be useful for increasing robustness.

The Lateral geniculate nucleus (LGN), then modulates output signals from the retina using feedback signals from higher visual areas (Usrey & Alitto, 2015), to ensure better classification.

In the primary visual cortex (V1), neural processing is similar to a bank of Gabor filters (GFB) with multiple orientations, spatial frequencies, etc. (De Valois, Albrecht, et al., 1982; De Valois, William Yund, et al., 1982; Ringach, 2002). Due to its diversity, the GFB decomposes the signal into a large number of disentangled features. These features pass through either simple or complex cells. Simple cells have a linear response and discard irrelevant information. Complex cells have a non-linear response to detect more complex features (Vintch et al., 2015). It may be that features generated by GFB, simple cells, and complex cells are necessary for downstream areas to increase robustness (Dapello et al., 2020).

In V1, multiple layers of neurons are organized as cortical minicolumns. Each minicolumn receives inputs from the same receptive field and all minicolumns have the same receptive field *size*. Thus, each minicolumn is similar to a vector encoding features like pose, orientation, scale, etc. (Hinton, 1981; Hubel & Wiesel, 1963). This preserves the spatial relationships within an object, which is crucial for downstream object perception , and, thus, we suggest that it may increase robustness.

From V1 to V4, pooling takes place (Freeman & Simoncelli, 2011) causing an increase in receptive field sizes and hence creating different scales across the visual stream. Pooling may help to reduce dimensions by removing irrelevant information and creating abstractions of the object (Poggio et al., 2014), and, thus, we suggest that it may increase robustness.

Functionally, the entire cortex is known to encode the input signals in a sparse way, which improves the selectivity of the class-relevant features (Paiton et al., 2020), and, thus, we suggest that it may increase robustness.

Additionally, all neuronal processes always include stochasticity, and such stochasticity is thought to contribute to the generalization of the signals (Echeveste & Lengyel, 2018), and, thus, we suggest that it may increase robustness.

Among the works that take inspiration from early vision models to improve the robustness of DCNNs, we can first outline the study of **Reddy et al., 2020**, which implemented two **sampling** methods from biological vision. One is the non-uniform sampling of the retina, and the other is the multi-scale sampling (due to pooling) of the cortex. The authors implemented both methods similarly, as briefly illustrated in Fig. 3. S1. First, samples were sampled with either method. Then each sample was processed by a DCNN. The outputs of all the processed samples were averaged to obtain the final classification output. The authors trained the DCNN with these sampling methods to increase robustness.

Then, **Dapello et al., 2020** imitated **neuronal features of V1**. The authors prefixed the DCNN with their custom model of V1, called VOneBlock. It consists of the GFB with parameters picked from empirical distributions (De Valois, Albrecht, et al., 1982; De Valois, William Yund, et al., 1982; Ringach, 2002); simple cell linearity implemented with ReLU and complex cell non-linearity implemented with quadrature phase-pair spectral power (Carandini et al., 1997); neuronal stochasticity with Poisson distribution parameters obtained from primate V1 (Softky & Koch, 1993). The authors trained the DCNN prefixed with their VOneBlock to increase robustness.

## 3.3. Feedback

Throughout the cortical visual system, information is processed through feedforward and feedback connections (Pennartz et al., 2019). Through feedback connections, higher layer contextual information modulates the activation patterns in the lower layers. Thus, it is implementing strong long-range spatial dependencies, global information extraction, perceptual grouping (Kreiman & Serre, 2020), and recognition of challenging images (Kar et al., 2019; Kietzmann et al., 2019). Since feedback encourages the use of more global information, we suggest that it may increase robustness (Elsayed et al., 2018; Olshausen, 2013).

One possible way of implementing feedback is Predictive Coding (Aitchison & Lengyel, 2017), which proposes that the brain continuously updates itself based on the prediction error between the input signal and its prediction.

We note the following studies that build on bio-inspired feedback mechanisms to improve robustness. **Huang et al., 2020** used predictive coding to construct a DCNN with feedback. Feedback is implemented with a modified Deconvolutional Generative Model (DGM; Nguyen et al., 2019). DGM takes the output from layer $(l+1)$ of the DCNN, and then deconvolutes (transpose of convolution; reverse of convolution) to generate (predict) the output of layer $(l)$. Input images $h$, are processed by the feedforward DCNN to produce intermediate representations $z$ and predict the output label $y$. In the feedback pass, the predicted labels $y$ are used to reconstruct the intermediate representations, $z$, and then eventually to reconstruct the input as $\hat{h}$. As illustrated in Fig. 3. S4, the reconstruction $(\hat{h})$ of the input image $(h)$, intermediate latent representation $(z)$, and predicted output label $(y)$ are dynamically modulating each other through feedforward and feedback processes. The authors trained their DCNN+feedback model using classification and reconstruction based losses to increase robustness of the DCNN.

Then, Capsule Networks (CapsNets) by Sabour et al., 2017 implement cortical minicolumn-like structure and also a predictive coding based feedback process. A Capsule is a vector of neurons that captures object features as well as its instantiation parameters, such as pose, orientation, lighting, etc (Fig 3. S3). CapsNets implement feedback using Routing by Agreement. The weights between the capsules in layer $(l + 1)$ and capsules in layer $(l)$ are updated based on how well the lower layer capsule is able to predict the output of the higher layer capsule. This is done by finding the correlation (agreement) between the predictions of the higher and lower layer capsules. If the correlation is very high then the weight is increased (excitatory connection) else the weight is decreased to suppress irrelevant capsules (inhibitory connection). CapsNets use classification and reconstruction losses to update themselves.

**Qin et al., 2019** found that since CapsNets bear a very high resemblance to human representations (Sabour et al., 2017), it takes a very large amount of adversarial noise to misclassify them. When large adversarial noise is added to the image, it resembles the target class more than the original/ source class of the image (Santurkar et al., 2019). This causes the reconstructed image to resemble the target class more than the original class of the input. The authors used this discrepancy between the original and the reconstructed image for the successful detection of adversarial perturbations. For a given input sample $x$, the CapsNet reconstructs it using the predicted class information, as $x'$. Then the $L_2$ (euclidean) distance between them is found as $d(x, x')$. If this distance exceeds a preset threshold, then the input $x$ is said to be adversarial. Else, it is declared a normal image. The authors showed that their CapsNet based detection is successful in the detection of adversarial samples.

Finally, **Kim et al., 2020** implemented the entire visual system (i.e., retina, LGN, V1-V4, feedback and lateral inhibition, neuronal stochasticity, and sparse coding) as closely as possible. The authors then prefixed it to a DCNN to increase robustness.

## 3.4. Miscellaneous

In addition to early vision and feedback, other bio-inspired components have been studied recently. Actually, it has been argued that neurophysiological data itself may contain robust representations that are used in the human visual system. Thus, as a proof of concept, **Li et al., 2019** regularized DCNNs using neurophysiological data obtained from mice V1. The authors first measured the mouse V1 neuronal responses for each image and then computed the response similarity matrix for the image pairs. Then, the similarity matrix was used to regularize the DCNN to increase robustness.

In addition, sleep is thought to be significant for the retention of memory. The short-term memory content is transmitted to long-term memory when sleeping (Rasch & Born, 2013), i.e., memory consolidation. Memory consolidation (Rasch & Born, 2013) reduces overfitting and improves generalization (Lewis & Durrant, 2011; Wamsley et al., 2010; González et al., 2020; Wei et al., 2018), thus, we suggest it may increase robustness. **Tadros et al., 2020** thus implemented a sleep-like algorithm by using Spiking Neural Networks (SNNs; Diehl et al., 2015). The authors used SNNs and spike time-dependent plasticity (STDP; Song et al., 2000) to implement the sleep-like algorithm. They first trained the DCNN and then transformed it into an SNN. Then the training images were fed into the SNN to induce the reactivation of the neurons, which leads to STDP. STDP enables the weights of highly related neurons to be strengthened and weakly related neurons to be weakened, to increase robustness.

*Table 1. Summary of studies. We refer to each paper as S# instead of cross-referencing. 'Components' refers to the bio-inspired feature the authors used. 'Contribution' refers to the possible reasons for robustness.*

| Study | Main Author | Components | Contribution |
|-------|-------------|------------|--------------|
| S1 | Reddy et al., 2020 | Non-uniform and Multiscale sampling | High SNR, Scale and translation invariance |
| S2 | Dapello et al., 2020 | V1 neuronal features | Feature extraction and generalization |
| S3 | Qin et al., 2019 | Cortical minicolumn and Feedback | Object based representation |
| S4 | Huang et al., 2020 | Feedback (self consistency) | Predictive coding |
| S5 | Kim et al., 2020 | Anatomical features of visual stream and Feedback | Decorrelation and sparse coding |
| S6 | Li et al., 2019 | Neurophysiological data | Inductive bias |
| S7 | Tadros et al., 2019 | SNN + STDP | Feature abstraction |

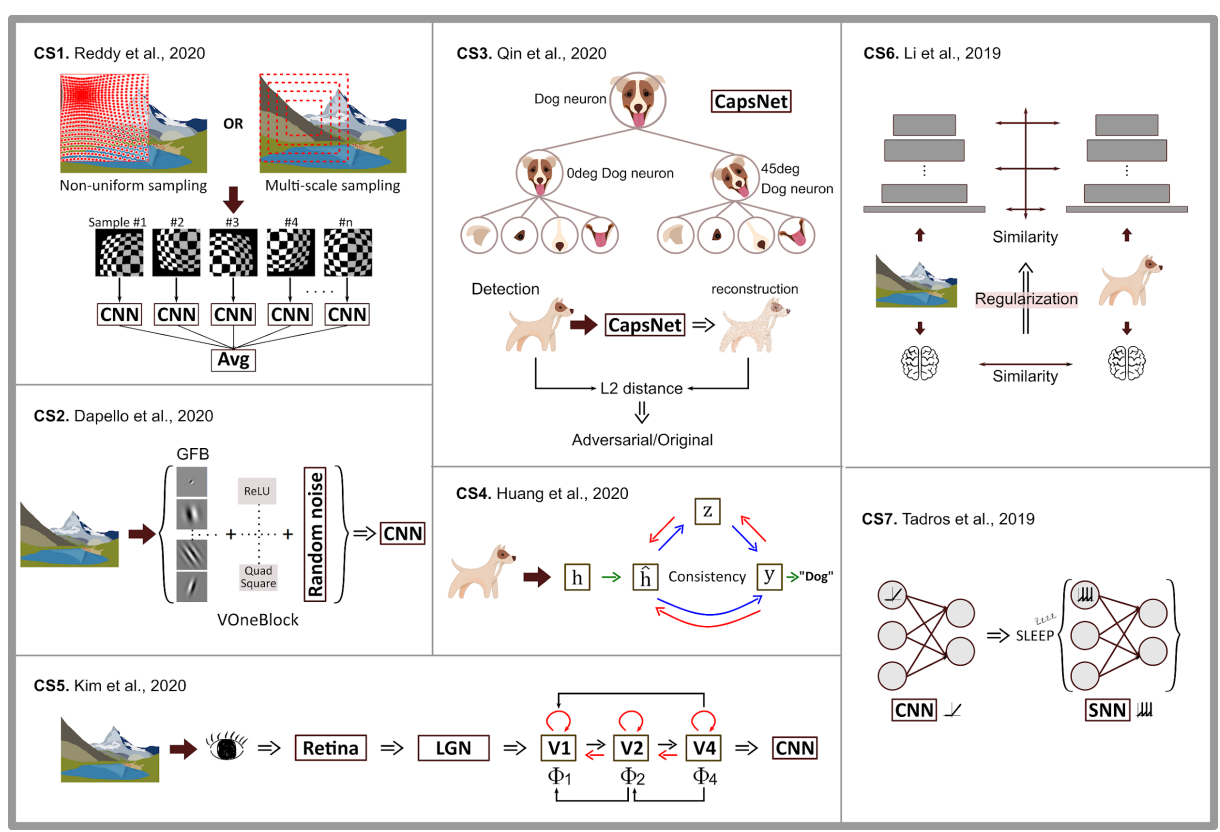

*Figure 3. Biologically inspired components. **S1**. Retinal non-uniform and multi-scaling sampling methods implemented by Reddy et al., 2020. Samples were sampled from an image using either non-uniform or multi-scale methods. A DCNN processes each sample and then averages the outputs to obtain the final classification prediction. **S2**. V1-like block, VOneBlock, prefixed on DCNN proposed by Dapello et al., 2020. VOneBlock consists of a Gaussian Filter Bank (GFB), linear and non-linear activation functions, and stochasticity. It pre-processes the input image and feeds the output into a DCNN. **S3**. Capsule Network (CapsNet) and the adversarial noise detection method suggested by Qin et al. (2020). Qin et al. (2020) detected the adversarial sample by using the $L_2$ (Euclidean) distance between the input image and the reconstructed image from CapsNet. **S4**. Recurrent model proposed by Huang et al., 2020. Input images h, are processed by the feedforward DCNN to produce intermediate representations z and predict the output label y. In the feedback pass, the predicted labels y are used to reconstruct z and then h. The reconstruction ($\hat{h}$) of the input image (h), the intermediate latent representation (z), and predicted output label (y) are dynamically modulating each other through feedforward and feedback processes. **S5**. Overall visual components implemented in the model proposed by Kim et al., 2020. Retina to V4 processes are implemented as a pre-processing module. The input images are first pre-processed by the module and then fed into the DCNN. **S6**. Mice V1 activity based DCNN regularization implemented by Li et al., 2019. The authors measured mouse V1 neuronal activation similarities given two different images and then used this similarity measure to regulate the activation patterns of DCNNs. **S7**. Sleep model implemented by Tadros et al., 2019. DCNNs are first trained in a standard way and then transformed to SNN, finally, the weights are consolidated by STDP.*

# 4. Analysis

We have seen how different components of biological vision can help to obtain more robustness against adversarial attacks. In this section, we analyze the evaluation of robustness as carried out by these studies (Table 1) and offer insights for future evaluation.

## 4.1. Evaluation

In Table 2, we summarize the list of attacks, metrics, datasets, and norms used for different works. As we can see, it is very difficult to compare across methods and understand the significance of the respective bio-inspired component (Table 1).

### 4.1.1 Experimental Settings

Evaluation of robustness is dependent on the dataset used to generate adversarial perturbations. The robustness of a DCNN on a more complex and realistic dataset like ImageNet (Deng et al., 2009) usually indicates the robustness of the DCNN on less complex datasets (Shafahi et al., 2020) like CIFAR10 (Krizhevsky, 2009). Datasets like ImageNet have more natural and realistic images, and hence more human vision-like features can be learned (Salman et al., 2020). Thus, evaluation of the robustness of DCNN on more complex datasets is a stronger result than on less complex datasets. S1, 2, and 5 have evaluated the robustness of their models with the more complex ImageNet dataset, while the other studies show their robustness on less complex datasets (Table 2). Furthermore, for evaluating a models' robustness, adversarial perturbations should be created w.r.t. the model. All studies (except S5) indeed attack their proposed models.

Then, the choice of $\epsilon$ values has a crucial role in analyzing the performance of the different proposals. In S2, for example, accuracy is hard to evaluate since accuracy was averaged across $\epsilon$ values and $L_p$ norms (Table 2). While integrating adversarial accuracy over $\epsilon$ values is acceptable (Bouniot et al., 2021), it becomes problematic when averaged over two values $\{\frac{1}{1020}, \frac{1}{255}\}$.

Secondly, as mentioned in Sec 2.1, each $L_p$ norm has a very different meaning (Tramèr & Boneh,

2019). Therefore, for a fair comparison with other studies, we obtained the non-averaged adversarial accuracy for the main model from the supplementary material provided by the authors. Further, we re-evaluated the robustness of the base models used in the paper without any averaging and presented them in Fig. 4. We observe that the authors reported an average adversarial accuracy (across $\epsilon$ values and $L_p$ norms) of 51.1% for their method and 52.3% for Adversarially trained DCNN, but in actuality with the $L_\infty$ norm, the adversarial accuracy differences ($\Delta acc$ = adv. acc. of their model - adv. acc. of AT DCNN) were different for different $\epsilon$ values ($\epsilon = \frac{1}{1020}, \Delta acc = 1.1\%$; $\epsilon = \frac{1}{255}; \Delta acc = -25\%$; $\epsilon = \frac{4}{255}; \Delta acc = -33.58\%$).

*Table 2. Evaluation ' * ' indicates a customized version of the adversarial attack created by the authors. AA indicates Adversarial accuracy, UDR (undetected rate) refers to the fraction of adversarial samples undetected by the method from all the generated samples, MD Indicates the median noise distortion. All the above metrics are calculated for a given value of $\epsilon$ and a given $L_p$. ' ✛ ' indicates adversarial accuracy that had been averaged over various $L_p$ norms and $\epsilon$ values as mentioned by the authors of S2.*

| Study | Image Dataset | $L_p$ norm | Metric | Attacks used |
|---|---|---|---|---|
| S1 | CIFAR10, ImageNet | *1, 2, ∞* | AA | PGD, FGSM, PGD ADAM, L2 CW, DBA, PGD BPDA |
| S2 | ImageNet-Val | *1, 2, ∞* | ✛ | PGD+MC |
| S3 | MNIST, F-MNIST, SVHN | *2, ∞* | UDR | PGD, PGD-R*, CW, BIM |
| S4 | F-MNIST, CIFAR10 | ∞ | AA | PGD, PGD*, SPSA |
| S5 | ImageNet-Val | x | AA | PGD |
| S6 | CIFAR10 | 0, 1, 2, ∞ | MD | PGD, DBA, etc following list of attacks by Brendel et al., 2019 |
| S7 | Patch, MNIST | 2 | MD | DeepFool, JSMA, DBA, FGSM |

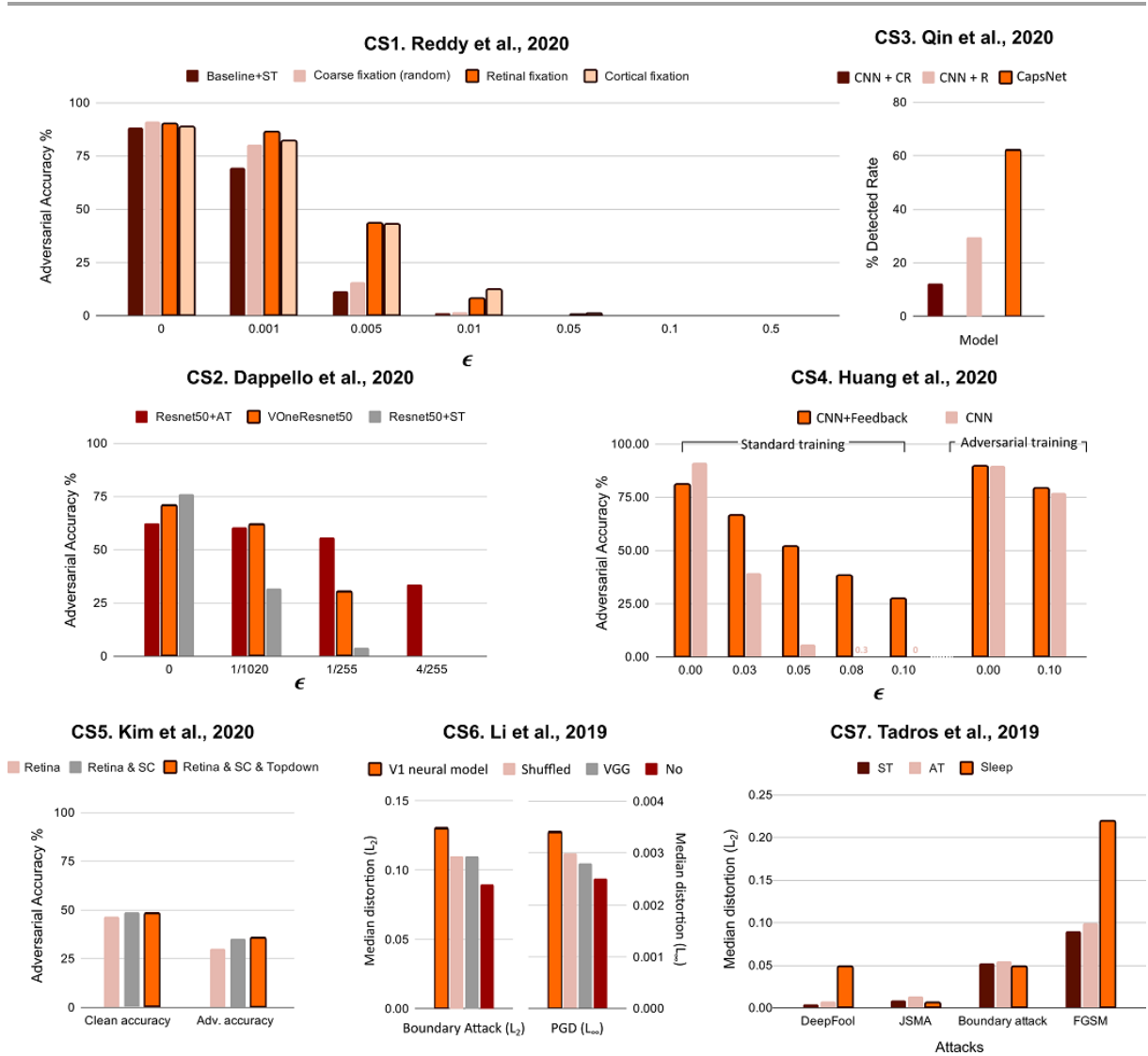

*Figure 4. Evaluation summary. The orange bar indicates robustness w.r.t. the metrics used (Table 2) from the proposed contributions, except for S3 where we used (1-UDR) metric for ease of understanding. For all studies, the **higher** the orange bar is, the **better** is the adversarial robustness of the respective method. From Table 2, we see that many of the studies have worked with multiple datasets and norms. For the sake of easier understanding, we chose to only present the results for the most complex image dataset used (Complexity is approximately: ImageNet (Deng et al., 2009) > CIFAR10 (Krizhevsky, 2009) > SVHN (Netzer et al., 2011) > F-MNIST (Xiao et al., 2017) > MNIST (LeCun & Cortes, 2010)). We also chose to present the results of the common $L_\infty$norm for all works with the exception of S7, which only tested on the $L_2$norm. Additionally, for S4, we present the results from F-MNIST instead of CIFAR10 since it is the only dataset from the paper for which standard and adversarially trained feedback models are compared. Similarly, for S1, we chose to present the results for ImageNet10, a 10 class subset of ImageNet, since it had a more comprehensive evaluation. For S2 and 5, the ImageNet dataset was used. For S3, we use SVHN; for S6 we use CIFAR10; and for S7, we use MNIST since it was the most complex image dataset used.*

Currently, most of the adversarial attack methods used for evaluation are iterative (Madry et al., 2017; Athalye et al., 2018; Carlini et al., 2019). Madry et al., (2017) showed that with a small increase of the number of iterations (from 7 to 20), adversarial accuracy deteriorates significantly (~ 5%; Table 2 of Madry et al., 2017). This indicates that one should always have *enough iterations* to create the adversarial samples.

A simple way of finding the required number of iterations for an attack is to pit the *number of iterations against adversarial accuracy*. After the required number of iterations, the *adversarial accuracy* will saturate close to 0, indicating that more iterations after this point do not result in better adversarial samples. Only S2 did such an analysis to decide the number of iterations (Table 3). Since this analysis requires large computational resources, it is not easy to do. The alternative is to just give a sufficiently large number of iterations as suggested by Madry et al., (2017) and carried out by S3.

*Table 3. Worst-case evaluation parameters: For each study, we picked the most complex image dataset and the strongest attack. All attacks in the table are for $L_\infty$ **norm**. ' ? ' Indicates values missing in the original paper. We excluded S6 and 7 from the table since the authors used the mean distortion metric to find the minimum perturbation needed to cause a misclassification. For doing so, by definition, a large number of iterations are needed (1000 or more).*

| Study | Dataset | Attack | $\epsilon$ ranges | #iterations | #iterations vs Adv. Acc analysis |
|---|---|---|---|---|---|
| S1 | ImageNet10 | PGD | {0.001, 0.005, 0.01, 0.02, 0.05, 0.1, 0.5} | 5 or 20 | x |
| S2 | ImageNet-Val | PGD-MC | $\{\frac{1}{1020}, \frac{1}{255}\}$ | 64 | ✓ |
| S3 | SVHN | PGD-R* | 0.1 | 200 | Not required |
| S4 | CIFAR10 | PGD* | 8/255 | 7 | x |
| S5 | ImageNet-Val | PGD | ? | ? | ? |

Further, for S1 and 2, we can see that for the Imagenet dataset (ImageNet10 is a 10 class subset of ImageNet), and the $L_\infty$ norm based attacks, a very different range of $\epsilon$ values are used for evaluation. For each dataset and $L_p$ norm, the values of $\epsilon$ are commonly agreed upon to be imperceptible (Madry et al., 2017; Bouniot et al., 2021). For example for ImageNet, the $L_\infty$ norm perturbations are tested with $\epsilon \in \{\frac{4}{255}, \frac{8}{255}, \frac{16}{255}\}$. Note that Elsayed et al. (2018) showed that $L_\infty$ norm perturbations up to $\epsilon \approx \frac{32}{255}$ are imperceptible to humans, which is much larger than the values tested by S2.

### 4.1.2 Understanding the bio-inspired components

**Component analysis** (Sec. 3.1)**.** All the studies did the component analysis by freezing each component or alternating the bio-inspired component with a simpler variant (Table 4).

S2 used the Gabor filter parameters sampled from empirical distribution from primate brain (V1 Gabor filter bank), and used the parameters from a uniform distribution (GFB parameters chosen uniformly) as the baseline model. Also, S2 implemented stochasticity using Poisson noise, which is inspired by primate V1 neurophysiology. However, it remains unclear whether bio-inspired noises are any better than random noise. It might be interesting to replace the bio-inspired Poisson noise in the S2 model with random noise to see whether the Poisson noise, determined from empirical data, really matters (Dhillon et al., 2018; Fawzi et al., 2016; Rakin et al., 2018)

S6 used a pairwise similarity matrix based on V1 neuronal activations (Neural similarity) as a regularizer. Thus, S6 was tested against two baselines: a random shuffle of their proposed similarity matrix (Shuffled similarity) and a VGG based similarity matrix (VGG similarity).

**Validate components.** For S3, 4, 5, and 7, since these studies mainly propose concepts like feedback and sleep for improving robustness, it is hard to find a simpler version of their model to be compared with. The baseline for such studies is the DCNN itself and all seven studies made this comparison.

S3 and 4 visualized the reconstructions using their feedback processes (Table 4). S4 also used Grad-CAM (Selvaraju et al., 2020) visualizations and showed that with more iterations of feedback, their model learned to extract better features from the images. S7 verified memory consolidation by weight visualization of the output layer weights, showing that indeed the connections with weak weights were pruned, and those with stronger weights were strengthened by their method.

*Table 4. Summary of the component analysis. Bio-inspired components of each study are listed in the 'Components' column. Under the 'Compared with' column, we list the baseline comparison models, which can either be the full model removing the corresponding component (ablated model) or a simplified model. Under*

*the 'Visualization' column, we listed the visualization method if the functionalities of the component were visualized.*

| Study | Components | Compared with | Visualization |
|---|---|---|---|
| S1 | Non-uniform Samp. | Uniform Sampling | - |
| | Multi-scale samp. | Single scale sampling | - |
| S2 | V1 Gabor filter bank | GFB parameters chosen uniformly | - |
| | Simple cell | Removal | - |
| | Complex cell | Removal | - |
| | Stochasticity | Removal & Random noise | - |
| S3 | Feedback | Removal: baseline DCNN | Reconstruction |
| S4 | Feedback | Removal: baseline DCNN | Reconstruction, Grad-CAM |
| S5 | Retinal structure | Removal: baseline DCNN | - |
| | Sparse coding | Removal | - |
| | Feedback | Removal | Reconstruction, weight visualization |
| S6 | Neural similarity | Shuffled and VGG similarities | - |
| S7 | SNN | Removal: baseline DCNN | Weight visualization |

***Attacks for Feedback.*** Feedback based models usually relay the feedback output from each iteration to the feedforward sweep of the next iteration. These types of models, when unrolled across iterations, are equivalent to extremely deep feedforward models (Kreiman & Serre, 2020). Athalye et al., 2018 have shown that for these deep feedforward models, there is a chance that gradients vanish or explode while calculating the adversarial perturbation (Eq. 1). This is because existing adversarial attack algorithms are not optimized for such deep architectures (taking too many iterations). The authors of both S3 and 4 designed their own variants of the adversarial attacks to account for the vanishing/exploding gradient effect. The robustness evaluation results in S5 are unfortunately incomplete from that point of view.

### 4.1.3 Evaluation Completeness

**Check for gradient masking.** As stated in Sec.2.2.2, gradient masking gives a false sense of robustness. As seen in Table 5, most of the studies checked for gradient masking specially the ones which have potential components that could cause gradient masking.

*Table 5. Checklist for gradient masking and relevant attacks. 'Components' refers to the feature that possibly causes gradient masking. The numbering in the 'Checklist' column corresponds to the checklist for gradient masking provided in Section 2.2.2. 'Relevant attacks' column refers to similar adversarial attacks, which were used to break gradient masking based defenses having similar components which have been listed in the column, titled 'Similar defenses'. 'Compared to AT' column refers to the studies which have compared themselves to Adversarial Training. For all columns, untick '✓' represents the studies that did respective columns' analysis and ' x ' represents the opposite; ' - ' represents not relevant to the study. Since all studies, except S3, change decision boundaries by training hence need to be compared to Adversarial Training (AT).*

| Study | Components | Checklist | Relevant attacks | Similar defense | Compared to AT |
|---|---|---|---|---|---|
| S1 | Downsampling | 1, 2, 4, 5 | PGD+BPDA | Guo et al., 2018 | ✓Only for CIFAR10 but trained on different $\epsilon$ |
| S2 | Stochasticity | 2 | PGD-MC | Dhillon et al., 2018 | ✓ but trained on different $\epsilon$ |
| S3 | - | 4 | - | - | - |
| S4 | - | 2, 3 | - | - | ✓ |
| S5 | Stochasticity | x | x | Dhillon et al., 2018 | x |
| S6 | - | 6 | - | - | x |
| S7 | - | 2, 6 | - | - | ✓ |

As seen in Table 5, S2 acknowledges that stochasticity based defenses (Dhillon et al., 2018) have been broken in the past (Carlini et al., 2019). By finding the averaged value of the gradient for each attack over many random trials, the stochastic nature of the defense can be broken (PGD-MC; Athalye et al., 2018). Similarly, S1 recognizes that defenses with non-differentiable operations like downsampling and Gaussian blur (Guo et al., 2018) have been broken by utilizing a differentiable approximation of these operations (PGD+BPDA; Athalye et al., 2018). Both S1 and S2 showed that their models were extremely robust even when tested with the PGD-MC and PGD-BPDA attacks, respectively.

***Comparison to Adversarial Training.*** Finally, for defenses, which aim to make DCNNs robust to adversarial perturbations by modifying their decision boundary, the gold standard is to compare their robustness with an Adversarially Trained DCNN. Except for S6 and 7, this was true for all studies (Table 5; Fig. 4).

Further, Adversarial Training depends on the attack used to generate the adversarial samples, which depend on the $\epsilon$ parameter and the $L_p$ norm. The robustness of Adversarial training depends heavily on the $\epsilon$ value (Madry et al., 2017). For a fair comparison of methods, DCNNs should, hence, be adversarially trained and tested on the same $\epsilon$ values. However, S1 and 2 compare their model with an adversarially trained DCNN with a fixed $\epsilon_{train}$ value whereas the testing was carried out with much lower $\epsilon_{test}$ values (S1 used $\epsilon_{train}$= 8/255 and $\epsilon_{test}$: {0.001, 0.005, 0.01, 0.05, 0.1, 0.5}; S2 used $\epsilon_{train}$= 4/255 and $\epsilon_{test}$: $\{\frac{1}{1020}, \frac{1}{255}\}$).

## 4.2. Modeling Insights

Many models have proposed that adding biological components can improve robustness to adversarial attacks. However, as we have seen in the last section, whether these attempts were successful cannot easily be judged because often evaluation is hard to carry out. Here, we provide some useful insight gained from these studies and insights for future research directions.

***Signal-to-noise ratio (SNR).*** It seems that S1 and S5 improve robustness by increasing the SNR, which is definitely a viable method. We suggest that this is the case because a high SNR may increase the relevant signals and discard the irrelevant signals (similar to Xie et al., 2019).

***Stochasticity.*** S2 and 5 used stochasticity to improve generalization hence increasing robustness. While in biological systems, all neurons are noisy, S2 and 5 implement stochasticity in only one layer of the DCNN. It would be interesting to see how multiple layers of stochasticity would affect the robustness of the DCNN (Rakin et al., 2018).

***Feedback process for the inference and error correction.*** As seen in Figure 4, S3 and 4 are extremely robust to adversarial perturbations by using feedback connections. Furthermore, with multiple runs of feedforward and feedback processing, it would be possible to learn more abstract and global features capable of generalizing, as shown by Doerig, Bornet, et al., 2020; Doerig, Schmittwilken, et al., 2020 and Kreiman & Serre, 2020, which should in turn increase robustness to adversarial attacks.

Currently, the only limiting factor for running multiple feedback loops in larger models is the large computational requirement, mainly because recurrent models are hard to parallelize for current GPUs. In the future, it may be possible to implement more recurrent long-short range connections like S4 along with inhibitory and excitatory connections as it is done in S3 (Zhao & Huang, 2019).

***Other Mechanisms.*** In S7, a sleep mechanism was implemented using SNN and STDP (memory consolidation). S7 provides good evidence that a well-established cognitive-behavioral component can help to increase robustness. Therefore, we expect that further cognitive features can increase robustness.

S6 showed the possibility of using neurophysiological data for robustness. S6 visualized its proposed similarity matrix to show the V1 neuronal similarity between different image pairs. However, the authors did this only for images without any adversarial noise. It would be interesting if the same similarity matrix could be obtained for a small validation set of adversarial samples to show that the current approach of S6 is already capable of having similar neural activations for adversarial images (like humans).

In summary, the features that improve adversarial robustness are as follows: 1) Increasing the signal-to-noise ratio, as in S1, 2, and 5. 2) Generalization by using stochasticity, as in S2 and 5. 3) Force the network to learn better representations, as in S3, 4, 6, and 7.

# 5. Discussion

Adversarial Attacks pose a serious problem for state-of-the-art DCNNs in general. In particular, if there are no easy fixes found, it is clear that DCNNs cannot be good models for the human visual system. For this reason, many researchers have tried to defend DCNNs against Adversarial Attacks by using inspirations from the human visual system. Here, we have reviewed the most important recent approaches and found that indeed some bio-inspired components that reproduce stochasticity and feedback phenomena increase robustness. However, due to the immense diversity of evaluation metrics, parameters, or datasets chosen for performance evaluation, it is very hard to judge if a certain component of biological vision is actually useful for increasing robustness (Sec. 4.2). We have reviewed evaluation criteria that could help standardize the evaluation for bio-inspired components. Finally, we summarized insights for future research directions towards designing more robust DCNNs, and generally better understanding of human vision and DCNNs in the presence of perturbations of data.

While we believe such an approach could help the robustness study of DCNN, a question remains.

**One man's trash, another man's treasure?** Humans primarily use global rather than local visual information. Hence, it may be attractive to encourage the use of more global information for DCNNs through bio-inspired components. However, it is an open question whether including global information can lead to side effects, such as susceptibility to illusions (Watanabe et al., 2018; Pang et al., 2021; Lonnqvist et al., 2021), which, if true, would lead to an explanation of why human vision is often non-veridical.

Through our paper, we have seen how components from the human visual system can help DCNNs to become more robust to adversarial perturbations. However, it is also possible to gain insights from DCNNs to explain phenomena of the human visual system, similar to reinforcement learning (Hassabis et al., 2017). We believe that studying the high vulnerability of current DCNNs can help to explain the adversarial robustness of humans. For example, both Dhillon et al., 2018 and Rakin et al., 2018 showed that random noise increases the robustness of DCNNs. They proposed that this may be due to the generalization properties induced by random noise. This, in turn, tells us that neuronal stochasticity may make humans robust to adversarial attacks. We suggest that such inferences made from the adversarial robustness of DCNNs may, in fact, give us more understanding of how different components of the human visual system function.

In summary, current DCCNs are not good models of the human visual system because they are vulnerable to adversarial attacks. As we have shown, there is not yet a single, simple fix to this vulnerability, as, for example, proposed by Firestone (2019). However, adding many features of the human visual system into DCNNs, in particular feedback and noise, clearly increases the robustness of DCNNs and thus helping bridge the gap between DCNNs and human visual processing. We further argue that we can learn important issues about human visual processing by studying why DCCNs fail to be robust, i.e., we may learn from imperfect models as much as having a perfect model (Lonnqvist et al., 2021). However, commonly agreed criteria on how to evaluate methods to increase robustness are crucial for all these efforts. Our review is a contributive step towards this goal.

## Acknowledgement

HM is financed by an interdisciplinary EPFL i-SV grant between Profs. Frossard and Herzog. OHC is financed by the Swiss National Science Foundation (SNF) 320030_176153 "Basics of visual processing: from elements to figures".

# References

Aitchison, L., & Lengyel, M. (2017). With or without you: Predictive coding and Bayesian inference in the brain. *Current Opinion in Neurobiology*, *46*, 219–227. https://doi.org/10.1016/j.conb.2017.08.010

Akhtar, N., & Mian, A. (2018). Threat of adversarial attacks on deep learning in computer vision: A survey. *IEEE Access*, *6*, 14410–14430.

Athalye, A., Carlini, N., & Wagner, D. (2018a). Obfuscated Gradients Give a False Sense of Security: Circumventing Defenses to Adversarial Examples. *ArXiv:1802.00420 [Cs]*. http://arxiv.org/abs/1802.00420

Athalye, A., Carlini, N., & Wagner, D. A. (2018b). Obfuscated Gradients Give a False Sense of Security: Circumventing Defenses to Adversarial Examples. *CoRR*, *abs/1802.00420*. http://arxiv.org/abs/1802.00420

Bakshi, A., & Ghosh, K. (2017). Chapter 26—A Neural Model of Attention and Feedback for Computing Perceived Brightness in Vision. In P. Samui, S. Sekhar, & V. E. Balas (Eds.), *Handbook of Neural Computation* (pp. 487–513). Academic Press. https://doi.org/10.1016/B978-0-12-811318-9.00026-0

Bochkovskiy, A., Wang, C.-Y., & Liao, H.-Y. M. (2020). YOLOv4: Optimal Speed and Accuracy of Object Detection. *ArXiv:2004.10934 [Cs, Eess]*. http://arxiv.org/abs/2004.10934

Bouniot, Q., Audigier, R., & Loesch, A. (2021). Optimal Transport as a Defense Against Adversarial Attacks. *ArXiv:2102.03156 [Cs, Stat]*. http://arxiv.org/abs/2102.03156

Brendel, W., & Bethge, M. (2017). Comment on "Biologically inspired protection of deep networks from adversarial attacks." *ArXiv:1704.01547 [Cs, q-Bio, Stat]*. http://arxiv.org/abs/1704.01547

Brendel, W., Rauber, J., & Bethge, M. (2018). Decision-Based Adversarial Attacks: Reliable Attacks Against Black-Box Machine Learning Models. *ArXiv:1712.04248 [Cs, Stat]*. http://arxiv.org/abs/1712.04248

Brendel, W., Rauber, J., Kümmerer, M., Ustyuzhaninov, I., & Bethge, M. (2019). Accurate, reliable and fast robustness evaluation. *ArXiv:1907.01003 [Cs, Stat]*. http://arxiv.org/abs/1907.01003

Carandini, M., Heeger, D. J., & Movshon, J. A. (1997). Linearity and normalization in simple cells of the macaque primary visual cortex. *The Journal of Neuroscience: The Official Journal of the Society for Neuroscience*, *17*(21), 8621–8644.

Carlini, N., Athalye, A., Papernot, N., Brendel, W., Rauber, J., Tsipras, D., Goodfellow, I., Madry, A., & Kurakin, A. (2019). On Evaluating Adversarial Robustness. *ArXiv:1902.06705 [Cs, Stat]*. http://arxiv.org/abs/1902.06705

Carlini, N., & Wagner, D. (2017). Towards evaluating the robustness of neural networks. *2017 IEEE Symposium on Security and Privacy (SP)*, 39–57.

Chakraborty, A., Alam, M., Dey, V., Chattopadhyay, A., & Mukhopadhyay, D. (2018). Adversarial Attacks and Defences: A Survey. *ArXiv:1810.00069 [Cs, Stat]*. http://arxiv.org/abs/1810.00069

Choi, Y., Uh, Y., Yoo, J., & Ha, J.-W. (2020). Stargan v2: Diverse image synthesis for multiple domains. *Proceedings of the IEEE/CVF Conference on Computer Vision and Pattern Recognition*, 8188–8197.

Choksi, B., Mozafari, M., O'May, C. B., Ador, B., Alamia, A., & VanRullen, R. (2020, October 9). *Brain-inspired predictive coding dynamics improve the robustness of deep neural networks*. NeurIPS 2020 Workshop SVRHM. https://openreview.net/forum?id=q1o2mWaOssG

Croce, F., Andriushchenko, M., Sehwag, V., Flammarion, N., Chiang, M., Mittal, P., & Hein, M. (2020). RobustBench: A standardized adversarial robustness benchmark. *ArXiv:2010.09670 [Cs, Stat]*. http://arxiv.org/abs/2010.09670

Dapello, J., Marques, T., Schrimpf, M., Geiger, F., Cox, D. D., & DiCarlo, J. J. (2020). *Simulating a Primary Visual Cortex at the Front of CNNs Improves Robustness to Image Perturbations* [Preprint]. Neuroscience. https://doi.org/10.1101/2020.06.16.154542

De Valois, R. L., Albrecht, D. G., & Thorell, L. G. (1982). Spatial frequency selectivity of cells in macaque visual cortex. *Vision Research*, *22*(5), 545–559. https://doi.org/10.1016/0042-6989(82)90113-4

De Valois, R. L., William Yund, E., & Hepler, N. (1982). The orientation and direction selectivity of cells in macaque visual cortex. *Vision Research*, *22*(5), 531–544. https://doi.org/10.1016/0042-6989(82)90112-2

Deng, J., Dong, W., Socher, R., Li, L., Kai Li, & Li Fei-Fei. (2009). ImageNet: A large-scale hierarchical image database. *2009 IEEE Conference on Computer Vision and Pattern Recognition*, 248–255. https://doi.org/10.1109/CVPR.2009.5206848

Dhillon, G. S., Azizzadenesheli, K., Lipton, Z. C., Bernstein, J. D., Kossaifi, J., Khanna, A., & Anandkumar, A. (2018, February 15). *Stochastic Activation Pruning for Robust Adversarial Defense*. International Conference on Learning Representations. https://openreview.net/forum?id=H1uR4GZRZ

Diehl, P. U., Neil, D., Binas, J., Cook, M., Liu, S., & Pfeiffer, M. (2015). Fast-classifying, high-accuracy spiking deep networks through weight and threshold balancing. *2015 International Joint Conference on Neural Networks (IJCNN)*, 1–8. https://doi.org/10.1109/IJCNN.2015.7280696

Doerig, A., Bornet, A., Choung, O. H., & Herzog, M. H. (2020). Crowding reveals fundamental differences in local vs. Global processing in humans and machines. *Vision Research*, *167*, 39–45.

Doerig, A., Schmittwilken, L., Sayim, B., Manassi, M., & Herzog, M. H. (2020). Capsule networks as recurrent models of grouping and segmentation. *PLoS Computational Biology*, *16*(7), e1008017.

Echeveste, R., & Lengyel, M. (2018). The redemption of noise: Inference with neural populations. *Trends in Neurosciences*, *41*(11), 767–770.

Elsayed, G., Shankar, S., Cheung, B., Papernot, N., Kurakin, A., Goodfellow, I., & Sohl-Dickstein, J. (2018). Adversarial examples that fool both computer vision and time-limited humans. *Advances in Neural Information Processing Systems*, 3910–3920.

Engstrom, L., Ilyas, A., Santurkar, S., Tsipras, D., Tran, B., & Madry, A. (2019). Adversarial Robustness as a Prior for Learned Representations. *ArXiv:1906.00945 [Cs, Stat]*. http://

arxiv.org/abs/1906.00945

Fawzi, A., Moosavi-Dezfooli, S.-M., & Frossard, P. (2016). Robustness of classifiers: From adversarial to random noise. *Advances in Neural Information Processing Systems*, 1632–1640.

Fezza, S. A., Bakhti, Y., Hamidouche, W., & Déforges, O. (2019). Perceptual Evaluation of Adversarial Attacks for CNN-based Image Classification. *ArXiv:1906.00204 [Cs, Eess, Stat]*. http://arxiv.org/abs/1906.00204

Freeman, J., & Simoncelli, E. P. (2011). Metamers of the ventral stream. *Nature Neuroscience*, *14*(9), 1195–1201.

Geirhos, R., Meding, K., & Wichmann, F. A. (2020). Beyond accuracy: Quantifying trial-by-trial behaviour of CNNs and humans by measuring error consistency. *ArXiv:2006.16736 [Cs, q-Bio]*. http://arxiv.org/abs/2006.16736

González, O. C., Sokolov, Y., Krishnan, G. P., Delanois, J. E., & Bazhenov, M. (2020). Can sleep protect memories from catastrophic forgetting? *Elife*, *9*, e51005.

Guo, C., Rana, M., Cisse, M., & Maaten, L. van der. (2018, February 15). *Countering Adversarial Images using Input Transformations*. International Conference on Learning Representations. https://openreview.net/forum?id=SyJ7ClWCb

He, K., Gkioxari, G., Dollár, P., & Girshick, R. (2017). Mask r-cnn. *Proceedings of the IEEE International Conference on Computer Vision*, 2961–2969.

He, K., Zhang, X., Ren, S., & Sun, J. (2016). Deep residual learning for image recognition. *Proceedings of the IEEE Conference on Computer Vision and Pattern Recognition*, 770–778.

Hinton, G. F. (1981). A parallel computation that assigns canonical object-based frames of reference. *Proceedings of the 7th International Joint Conference on Artificial Intelligence - Volume 2*, 683–685.

Hosseini, H., Kannan, S., & Poovendran, R. (2019). Dropping Pixels for Adversarial Robustness. *ArXiv:1905.00180 [Cs, Stat]*. http://arxiv.org/abs/1905.00180

Huang, Y., Gornet, J., Dai, S., Yu, Z., Nguyen, T., Tsao, D., & Anandkumar, A. (2020). Neural Networks with Recurrent Generative Feedback. *Advances in Neural Information Processing Systems*, *33*.

Hubel, D. H., & Wiesel, T. N. (1963). Shape and arrangement of columns in cat's striate cortex. *The Journal of Physiology*, *165*, 559–568. https://doi.org/10.1113/jphysiol.1963.sp007079

Ilyas, A., Santurkar, S., Tsipras, D., Engstrom, L., Tran, B., & Madry, A. (2019). Adversarial examples are not bugs, they are features. *Advances in Neural Information Processing Systems*, 125–136.

Jordan, M., Manoj, N., Goel, S., & Dimakis, A. G. (2019). Quantifying Perceptual Distortion of Adversarial Examples. *ArXiv Preprint ArXiv:1902.08265*.

Kar, K., Kubilius, J., Schmidt, K., Issa, E. B., & DiCarlo, J. J. (2019). Evidence that recurrent circuits are critical to the ventral stream's execution of core object recognition behavior. *Nature Neuroscience*, *22*(6), 974–983. https://doi.org/10.1038/s41593-019-0392-5

Karpathy, A., & Fei-Fei, L. (2015). Deep visual-semantic alignments for generating image descriptions. *Proceedings of the IEEE Conference on Computer Vision and Pattern*

*Recognition*, 3128–3137.

Kaur, S., Cohen, J., & Lipton, Z. C. (2019). Are Perceptually-Aligned Gradients a General Property of Robust Classifiers? *ArXiv Preprint ArXiv:1910.08640*.

Kietzmann, T. C., Spoerer, C. J., Sörensen, L. K. A., Cichy, R. M., Hauk, O., & Kriegeskorte, N. (2019). Recurrence is required to capture the representational dynamics of the human visual system. *Proceedings of the National Academy of Sciences of the United States of America*, *116*(43), 21854–21863. https://doi.org/10.1073/pnas.1905544116

Kim, E., Rego, J., Watkins, Y., & Kenyon, G. T. (2020). Modeling Biological Immunity to Adversarial Examples. *2020 IEEE/CVF Conference on Computer Vision and Pattern Recognition (CVPR)*, 4665–4674. https://doi.org/10.1109/CVPR42600.2020.00472

Kiritani, T., & Ono, K. (2020). Recurrent Attention Model with Log-Polar Mapping is Robust against Adversarial Attacks. *ArXiv:2002.05388 [Cs]*. http://arxiv.org/abs/2002.05388

Kreiman, G., & Serre, T. (2020). Beyond the feedforward sweep: Feedback computations in the visual cortex. *Annals of the New York Academy of Sciences*, *1464*(1), 222–241. https://doi.org/10.1111/nyas.14320

Krizhevsky, A. (2009). *Learning multiple layers of features from tiny images*.

Kubilius, J., Schrimpf, M., Nayebi, A., Bear, D., Yamins, D. L. K., & DiCarlo, J. J. (2018). CORnet: Modeling the Neural Mechanisms of Core Object Recognition. *BioRxiv*, 408385. https://doi.org/10.1101/408385

Kurakin, A., Goodfellow, I., & Bengio, S. (2017). Adversarial examples in the physical world. *ArXiv:1607.02533 [Cs, Stat]*. http://arxiv.org/abs/1607.02533

Kwabena Patrick, M., Felix Adekoya, A., Abra Mighty, A., & Edward, B. Y. (2019). Capsule Networks – A survey. *Journal of King Saud University - Computer and Information Sciences*. https://doi.org/10.1016/j.jksuci.2019.09.014

Lewis, P. A., & Durrant, S. J. (2011). Overlapping memory replay during sleep builds cognitive schemata. *Trends in Cognitive Sciences*, *15*(8), 343–351. https://doi.org/10.1016/j.tics.2011.06.004

Li, Z., Brendel, W., Walker, E. Y., Cobos, E., Muhammad, T., Reimer, J., Bethge, M., Sinz, F. H., Pitkow, X., & Tolias, A. S. (2019). Learning From Brains How to Regularize Machines. *ArXiv:1911.05072 [Cs, q-Bio]*. http://arxiv.org/abs/1911.05072

Lonnqvist, B., Bornet, A., Choung, O. H., Doerig, A., & Herzog, M. H. (2021). *A comparative biology approach to CNN modeling of vision: A focus on differences, not similarities.*

Madry, A., Makelov, A., Schmidt, L., Tsipras, D., & Vladu, A. (2017). Towards deep learning models resistant to adversarial attacks. *ArXiv Preprint ArXiv:1706.06083*.

Marchisio, A., Nanfa, G., Khalid, F., Hanif, M. A., Martina, M., & Shafique, M. (2020). Is Spiking Secure? A Comparative Study on the Security Vulnerabilities of Spiking and Deep Neural Networks. *2020 International Joint Conference on Neural Networks (IJCNN)*, 1–8. https://doi.org/10.1109/IJCNN48605.2020.9207297

Moosavi-Dezfooli, S.-M., Fawzi, A., & Frossard, P. (2016). Deepfool: A simple and accurate method to fool deep neural networks. *Proceedings of the IEEE Conference on Computer Vision and Pattern Recognition*, 2574–2582.

Nayebi, A., & Ganguli, S. (2017). Biologically inspired protection of deep networks from adversarial attacks. *ArXiv:1703.09202 [Cs, q-Bio, Stat]*. http://arxiv.org/abs/1703.09202

Nguyen, T., Ho, N., Patel, A., Anandkumar, A., Jordan, M. I., & Baraniuk, R. G. (2019). A Bayesian Perspective of Convolutional Neural Networks through a Deconvolutional Generative Model. *ArXiv:1811.02657 [Cs, Stat]*. http://arxiv.org/abs/1811.02657

Olshausen, B. A. (2013). 20 Years of Learning About Vision: Questions Answered, Questions Unanswered, and Questions Not Yet Asked. In J. M. Bower (Ed.), *20 Years of Computational Neuroscience* (pp. 243–270). Springer New York. https://doi.org/10.1007/978-1-4614-1424-7_12

Ortiz-Jimenez, G., Modas, A., Moosavi-Dezfooli, S.-M., & Frossard, P. (2020). Optimism in the face of adversity: Understanding and improving deep learning through adversarial robustness. *ArXiv Preprint ArXiv:2010.09624*.

Paiton, D. M., Frye, C. G., Lundquist, S. Y., Bowen, J. D., Zarcone, R., & Olshausen, B. A. (2020). Selectivity and robustness of sparse coding networks. *Journal of Vision*, *20*(12), 10–10. https://doi.org/10.1167/jov.20.12.10

Pang, Z., O'May, C. B., Choksi, B., & VanRullen, R. (2021). Predictive coding feedback results in perceived illusory contours in a recurrent neural network. *ArXiv:2102.01955 [Cs, q-Bio]*. http://arxiv.org/abs/2102.01955

Papernot, N., McDaniel, P., Jha, S., Fredrikson, M., Celik, Z. B., & Swami, A. (2016). The limitations of deep learning in adversarial settings. *2016 IEEE European Symposium on Security and Privacy (EuroS&P)*, 372–387.

Pennartz, C. M. A., Dora, S., Muckli, L., & Lorteije, J. A. M. (2019). Towards a Unified View on Pathways and Functions of Neural Recurrent Processing. *Trends in Neurosciences*, *42*(9), 589–603. https://doi.org/10.1016/j.tins.2019.07.005

Poggio, T., Mutch, J., & Isik, L. (2014). Computational role of eccentricity dependent cortical magnification. *ArXiv:1406.1770 [Cs, q-Bio]*. http://arxiv.org/abs/1406.1770

Qin, Y., Frosst, N., Sabour, S., Raffel, C., Cottrell, G., & Hinton, G. (2019, September 25). *Detecting and Diagnosing Adversarial Images with Class-Conditional Capsule Reconstructions*. International Conference on Learning Representations. https://openreview.net/forum?id=Skgy464Kvr

Rakin, A. S., He, Z., & Fan, D. (2018). Parametric Noise Injection: Trainable Randomness to Improve Deep Neural Network Robustness against Adversarial Attack. *ArXiv:1811.09310 [Cs]*. http://arxiv.org/abs/1811.09310

Rasch, B., & Born, J. (2013). About sleep's role in memory. *Physiological Reviews*.

Rathbun, D. L., Warland, D. K., & Usrey, W. M. (2010). Spike timing and information transmission at retinogeniculate synapses. *Journal of Neuroscience*, *30*(41), 13558–13566.

Reddy, M. V., Banburski, A., Pant, N., & Poggio, T. (2020, June 29). Biologically Inspired Mechanisms for Adversarial Robustness. *ArXiv:2006.16427 [Cs, Stat]*. Conference on Neural Information Processing Systems. http://arxiv.org/abs/2006.16427

Ringach, D. L. (2002). Spatial structure and symmetry of simple-cell receptive fields in macaque primary visual cortex. *Journal of Neurophysiology*, *88*(1), 455–463. https://doi.org/10.1152/jn.2002.88.1.455

Rusak, E., Schott, L., Zimmermann, R. S., Bitterwolf, J., Bringmann, O., Bethge, M., & Brendel,

W. (2020). A simple way to make neural networks robust against diverse image corruptions. *ArXiv:2001.06057 [Cs, Stat]*. http://arxiv.org/abs/2001.06057

Sabour, S., Frosst, N., & Hinton, G. E. (2017). Dynamic Routing Between Capsules. *ArXiv:1710.09829 [Cs]*. http://arxiv.org/abs/1710.09829

Salman, H., Ilyas, A., Engstrom, L., Kapoor, A., & Madry, A. (2020). Do Adversarially Robust ImageNet Models Transfer Better? *ArXiv:2007.08489 [Cs, Stat]*. http://arxiv.org/abs/2007.08489

Santurkar, S., Ilyas, A., Tsipras, D., Engstrom, L., Tran, B., & Madry, A. (2019). Image synthesis with a single (robust) classifier. *Advances in Neural Information Processing Systems*, 1262–1273.

Schrimpf, M., Kubilius, J., Lee, M. J., Murty, N. A. R., Ajemian, R., & DiCarlo, J. J. (2020). Integrative Benchmarking to Advance Neurally Mechanistic Models of Human Intelligence. *Neuron*, *108*(3), 413–423. https://doi.org/10.1016/j.neuron.2020.07.040

Segal, I. Y., Giladi, C., Gedalin, M., Rucci, M., Ben-Tov, M., Kushinsky, Y., Mokeichev, A., & Segev, R. (2015). Decorrelation of retinal response to natural scenes by fixational eye movements. *Proceedings of the National Academy of Sciences*, *112*(10), 3110–3115. https://doi.org/10.1073/pnas.1412059112

Selvaraju, R. R., Cogswell, M., Das, A., Vedantam, R., Parikh, D., & Batra, D. (2020). Grad-CAM: Visual Explanations from Deep Networks via Gradient-based Localization. *International Journal of Computer Vision*, *128*(2), 336–359. https://doi.org/10.1007/s11263-019-01228-7

Shafahi, A., Saadatpanah, P., Zhu, C., Ghiasi, A., Studer, C., Jacobs, D., & Goldstein, T. (2020). Adversarially robust transfer learning. *ArXiv:1905.08232 [Cs, Stat]*. http://arxiv.org/abs/1905.08232

Sharif, M., Bhagavatula, S., Bauer, L., & Reiter, M. K. (2016). Accessorize to a crime: Real and stealthy attacks on state-of-the-art face recognition. *Proceedings of the 2016 ACM SIGSAC Conference on Computer and Communications Security*, 1528–1540.

Sincich, L. C., Horton, J. C., & Sharpee, T. O. (2009). Preserving Information in Neural Transmission. *Journal of Neuroscience*, *29*(19), 6207–6216. https://doi.org/10.1523/JNEUROSCI.3701-08.2009

Softky, W. R., & Koch, C. (1993). The highly irregular firing of cortical cells is inconsistent with temporal integration of random EPSPs. *Journal of Neuroscience*, *13*(1), 334–350. https://doi.org/10.1523/JNEUROSCI.13-01-00334.1993

Song, C., He, K., Lin, J., Wang, L., & Hopcroft, J. E. (2020). Robust Local Features for Improving the Generalization of Adversarial Training. *International Conference on Learning Representations*. https://openreview.net/forum?id=H1lZJpVFvr

Song, S., Miller, K. D., & Abbott, L. F. (2000). Competitive Hebbian learning through spike-timing-dependent synaptic plasticity. *Nature Neuroscience*, *3*(9), 919–926. https://doi.org/10.1038/78829

Szegedy, C., Zaremba, W., Sutskever, I., Bruna, J., Erhan, D., Goodfellow, I., & Fergus, R. (2014). Intriguing properties of neural networks. *ArXiv:1312.6199 [Cs]*. http://arxiv.org/abs/1312.6199

Tadros, T., Krishnan, G., Ramyaa, R., & Bazhenov, M. (2019, September 25). *Biologically*

*inspired sleep algorithm for increased generalization and adversarial robustness in deep neural networks*. International Conference on Learning Representations. https://openreview.net/forum?id=r1xGnA4Kvr

Tramèr, F., & Boneh, D. (2019). Adversarial Training and Robustness for Multiple Perturbations. *ArXiv:1904.13000 [Cs, Stat]*. http://arxiv.org/abs/1904.13000

Tramer, F., Carlini, N., Brendel, W., & Madry, A. (2020). On adaptive attacks to adversarial example defenses. *ArXiv Preprint ArXiv:2002.08347*.

Tsipras, D., Santurkar, S., Engstrom, L., Turner, A., & Madry, A. (2018). Robustness may be at odds with accuracy. *ArXiv Preprint ArXiv:1805.12152*.

Usrey, W. M., & Alitto, H. J. (2015). Visual Functions of the Thalamus. *Annual Review of Vision Science*, *1*(1), 351–371. https://doi.org/10.1146/annurev-vision-082114-035920

Utrera, F., Kravitz, E., Erichson, N. B., Khanna, R., & Mahoney, M. W. (2020). Adversarially-Trained Deep Nets Transfer Better. *ArXiv:2007.05869 [Cs, Stat]*. http://arxiv.org/abs/2007.05869

Vintch, B., Movshon, J. A., & Simoncelli, E. P. (2015). A Convolutional Subunit Model for Neuronal Responses in Macaque V1. *Journal of Neuroscience*, *35*(44), 14829–14841. https://doi.org/10.1523/JNEUROSCI.2815-13.2015

Wamsley, E. J., Tucker, M. A., Payne, J. D., & Stickgold, R. (2010). A brief nap is beneficial for human route-learning: The role of navigation experience and EEG spectral power. *Learning & Memory*, *17*(7), 332–336.

Wang, X., Hirsch, J. A., & Sommer, F. T. (2010). Recoding of sensory information across the retinothalamic synapse. *Journal of Neuroscience*, *30*(41), 13567–13577.

Watanabe, E., Kitaoka, A., Sakamoto, K., Yasugi, M., & Tanaka, K. (2018). Illusory Motion Reproduced by Deep Neural Networks Trained for Prediction. *Frontiers in Psychology*, *9*. https://doi.org/10.3389/fpsyg.2018.00345

Wei, Y., Krishnan, G. P., Komarov, M., & Bazhenov, M. (2018). Differential roles of sleep spindles and sleep slow oscillations in memory consolidation. *PLoS Computational Biology*, *14*(7), e1006322.

Xie, C., Wu, Y., Maaten, L. v d, Yuille, A. L., & He, K. (2019). Feature Denoising for Improving Adversarial Robustness. *2019 IEEE/CVF Conference on Computer Vision and Pattern Recognition (CVPR)*, 501–509. https://doi.org/10.1109/CVPR.2019.00059

Xie, Cihang, Tan, M., Gong, B., Wang, J., Yuille, A., & Le, Q. V. (2020). Adversarial Examples Improve Image Recognition. *ArXiv:1911.09665 [Cs]*. http://arxiv.org/abs/1911.09665

Zhang, T., & Zhu, Z. (2019). Interpreting adversarially trained convolutional neural networks. *ArXiv Preprint ArXiv:1905.09797*.

Zhang, Y., Foroosh, H., David, P., & Gong, B. (2019). CAMOU: Learning Physical Vehicle Camouflages to Adversarially Attack Detectors in the Wild. *International Conference on Learning Representations*. https://openreview.net/forum?id=SJgEl3A5tm

Zhao, L., & Huang, L. (2019). Exploring Dynamic Routing As A Pooling Layer. *2019 IEEE/CVF International Conference on Computer Vision Workshop (ICCVW)*, 738–742. https://doi.org/10.1109/ICCVW.2019.00095

Zoran, D., Chrzanowski, M., Huang, P.-S., Gowal, S., Mott, A., & Kohli, P. (2020). Towards

Robust Image Classification Using Sequential Attention Models. *2020 IEEE/CVF Conference on Computer Vision and Pattern Recognition (CVPR)*, 9480–9489. https://doi.org/10.1109/CVPR42600.2020.00950